\documentclass[letterpaper]{article} 
\usepackage{aaai25}  
\usepackage{times}  
\usepackage{helvet}  
\usepackage{courier}  
\usepackage[hyphens]{url}  
\usepackage{graphicx} 
\usepackage{natbib}  
\usepackage{caption} 
\usepackage{algorithm}
\usepackage{algorithmic}

\usepackage{times}  
\usepackage{helvet}  
\usepackage{courier}  
\usepackage[hyphens]{url}  
\usepackage{graphicx} 
\usepackage{amsmath}
\usepackage{amssymb}
\usepackage{enumitem}

\usepackage{graphicx}
\usepackage{subcaption}

\usepackage{natbib}  
\usepackage{caption} 
\usepackage{tabularray}

\usepackage{newfloat}
\usepackage{listings}
\DeclareCaptionStyle{ruled}{labelfont=normalfont,labelsep=colon,strut=off} 
\floatstyle{ruled}
\newfloat{listing}{tb}{lst}{}
\floatname{listing}{Listing}
\title{Time Series Foundational Models: Their Role in Anomaly \\Detection and Prediction}
\author{
  Chathurangi Shyalika, Harleen Kaur Bagga, Ahan Bhatt, Renjith Prasad,\\ Alaa Al Ghazo, Amit Sheth
}
\affiliations{
  Artificial Intelligence Institute, University of South Carolina, USA\\

}

\usepackage{bibentry}

\begin{document}
\maketitle
\begin{abstract}
Time series foundational models (TSFM) have gained prominence in time series forecasting, promising state-of-the-art performance across various applications. However, their application in anomaly detection and prediction remains underexplored, with growing concerns regarding their black-box nature, lack of interpretability and applicability. This paper critically evaluates the efficacy of TSFM in anomaly detection and prediction tasks. We systematically analyze TSFM across multiple datasets, including those characterized by the absence of discernible patterns, trends and seasonality. Our analysis shows that while TSFMs can be extended for anomaly detection and prediction, traditional statistical and deep learning models often match or outperform TSFM in these tasks. Additionally, TSFMs require high computational resources but fail to capture sequential dependencies effectively or improve performance in few-shot or zero-shot scenarios. \noindent The preprocessed datasets, codes to reproduce the results and supplementary materials are available at  \url{https://github.com/smtmnfg/TSFM}.
\end{abstract}

%

\vspace{-2mm}
\section{Introduction}
Foundational models (FMs), including large language models (LLM), have demonstrated efficacy in a variety of applications, with a particular emphasis on natural language processing (NLP) \cite{achiam2023gpt} and computer vision (CV) \cite{liu2024visual}. This has resulted in an increasing interest in the use of FMs for time series analysis \cite{zhang2024large}. The impressive capabilities of FMs and LLMs, including generalizability across domains, data efficiency, advanced reasoning and pattern recognition, multimodal knowledge integration and easy optimization, offer significant potential for enhancing time series forecasting without requiring per-task retraining from scratch \cite{jin2024position,jin2023time}. Recently, there has been a growing trend in developing foundational models specifically tailored for time series analysis.

However, applying FMs for time series analysis has seen several limitations. A significant concern is the ambiguity surrounding the datasets on which TSFMs are trained, which can lead to overestimated model performance and misleading evaluations. This issue becomes particularly problematic when the training data does not accurately represent the diversity of real-world scenarios, resulting in models that perform well in controlled environments but struggle to generalize to diverse, practical applications \cite{arjovsky2019invariant}. Moreover, data leakage is a critical issue in time series modeling, especially concerning TSFMs. If the temporal ordering of data is not strictly maintained or if features that will be known in the future are inadvertently included during training, the model can be influenced by information from the test set. Data leakage can produce artificially inflated performance metrics, giving false impressions of the model's robustness \cite{kaufman2012leakage, rosenblatt2024data}.

Additionally, while TSFMs may excel in forecasting tasks in some domains, their generalizability to other tasks, such as anomaly detection and prediction (prediction refers to identifying an anomalous event beforehand, while detection involves identifying it after it has occurred) \cite{shyalika2024ri2ap}, is limited. Anomaly detection and prediction involve identifying rare and unexpected patterns. This process fundamentally differs from the sequential prediction tasks for which TSFMs are designed. Consequently, TSFMs may either miss subtle anomalies or mistakenly classify normal variations as anomalies, leading to high false positive rates \cite{chandola2009anomaly}. Furthermore, TSFMs are resource-intensive, requiring substantial computational power and extensive labeled datasets for training. Obtaining such datasets is challenging due to the high costs and the difficulty of labeling time-series data, especially when anomalies are rare \cite{aggarwal2017outlier}. These constraints highlight the need for caution when applying TSFMs beyond their intended scope and underscore the importance of developing specialized models for specific tasks, such as anomaly detection and prediction. While some TSFMs, such as Moment \cite{goswami2024moment} and TimeGPT \cite{garza2023timegpt}, can perform anomaly detection, current TSFMs are not explicitly designed for anomaly prediction. Hence, our study adapted the standard next-time series forecasting approach for the anomaly prediction task. Ultimately, in this analysis, we selected models based on the criteria that they must be explicitly designed and trained for either anomaly detection or forecasting tasks.


Our research contributions are as follows:
\begin{itemize}
    \setlength{\itemsep}{0pt} 
    \setlength{\parskip}{0pt} 
    \setlength{\parsep}{0pt}  
    \item Comprehensive analysis and evaluation of TSFMs for anomaly detection and prediction across five time-series datasets.
    \item Benchmarking TSFMs against traditional statistical and deep learning approaches.
    \item Analysis of the effectiveness of fine-tuning TSFMs.
    \item Investigation of the computational intensity of TSFMs compared to benchmark approaches.
\end{itemize}






\begin{table*}[!ht]
\scriptsize
\centering
\label{tab:results_transposed}
\begin{tabular}{c|c|c|c|c|c} 
\hline
Foundational Model                                                          & TimeGPT                                                             & \begin{tabular}[c]{@{}c@{}}FPT \end{tabular}                 & Time-MOE (base)                                                        & MOIRAI (base)                                                       & Chronos (tiny)                                                           \\ 
\hline
Type                                                                        & \begin{tabular}[c]{@{}c@{}}Pre-training from \\scratch\end{tabular} & \begin{tabular}[c]{@{}c@{}}Adapting LLMs \\for time series\end{tabular}    & \begin{tabular}[c]{@{}c@{}}Pre-training from \\scratch\end{tabular} & \begin{tabular}[c]{@{}c@{}}Pre-training from \\scratch\end{tabular} & \begin{tabular}[c]{@{}c@{}}Adapting LLMs \\for time series\end{tabular}  \\ 
\hline
Max Context Length                                                          & Unknown                                                             & 384                                                                        & 4096                                                                & 5000                                                                & 512                                                                      \\ 
\hline
Max Model Size                                                              & Unknown                                                             & 3.2M                                                                       & 2.4B                                                                & 311M                                                                & 710M                                                                     \\ 
\hline
Architecture                                                                & Encoder-Decoder                                                     & Decoder-Only                                                               & Decoder-Only                                                        & Encoder-Only                                                        & Encoder-Decoder                                                          \\ 
\hline
\begin{tabular}[c]{@{}c@{}}Zero-shot detection / \\Forecasting\end{tabular} & \checkmark                                           & \checkmark                                                  & \checkmark                                           & \checkmark                                           & \checkmark                                                \\ 
\hline
Probabilistic Forecasting                                                   & \checkmark                                           & x                                                                          & x                                                                   & \checkmark                                           & \checkmark                                                \\ 
\hline
Anomaly Detection                                                           & \checkmark                                           & \checkmark                                                  & x                                                                   & x                                                                   & x                                                                        \\ 
\hline
Anomaly Prediction                                                          & x                                                                   & x                                                                          & x                                                                   & x                                                                   & x                                                                        \\ 
\hline
\begin{tabular}[c]{@{}c@{}}Pre-training Data \\(size)\end{tabular}          & Unknown (100B)                                                      & \begin{tabular}[c]{@{}c@{}}NLP pretrained\\transformer models\end{tabular} & Time-300B (300B)                                                    & LOTSA (27B)                                                         & 84B                                                                      \\ 
\hline
Fine-tuning                                                                 & \checkmark                                           & x                                                                          & \checkmark                                                                    & \checkmark                                           & \checkmark                                                \\ 
\hline
Open Source                                                                 & x                                                                   & \checkmark                                                  & \checkmark                                           & \checkmark                                           & \checkmark                                                \\ 
\hline
Input Token                                                                 & Patch                                                               & Patch                                                                      & Point                                                               & Patch                                                               & Point                                                                    \\
\hline
\end{tabular}
\caption{Overview of the Reference Time Series Foundational Models}
\end{table*}

\vspace{-4mm}
\section{Literature Review}

The research in this field can be divided into two primary approaches: pre-training foundational models from scratch for time series and adapting large language models for time series. Pre-training foundational models from scratch for time series represents an emerging area of research aimed at overcoming the unique challenges inherent in time series data, where limitations in scale and variability have historically hindered the development of robust, generalized models. Foundational models, initially successful in NLP and computer vision through large-scale pre-training, have shown impressive zero-shot and few-shot learning capabilities, often outperforming task-specific models. However, as anticipation grows for similar foundational models tailored to time series, recent efforts like ForecastPFN \cite{dooley2024forecastpfn}, TimeGPT \cite{garza2023timegpt} mark pioneering steps towards advancing time series analysis with models capable of capturing the unique temporal dynamics and patterns inherent in this domain. Adapting LLMs for time series analysis involves leveraging their pre-trained capabilities for various downstream tasks, focusing on effectiveness, efficiency and explainability. Two main adaptation paradigms—embedding-visible \cite{jin2023time, zhou2023one} and textual-visible \cite{zhang2023large, liu2023large, xue2023promptcast} are inspired by NLP techniques, differing primarily in input-output approaches and how time series data is integrated. Beyond forecasting, LLMs can also serve as enhancers, data generators and explainers, expanding their utility across diverse time series applications.

The mainstream downstream time series tasks include classification, forecasting, imputation and anomaly detection, each addressing key aspects of temporal data analysis. Among these, anomaly detection and prediction, within the realm of time series foundation models, has been the least explored and remains particularly challenging due to several factors. The rarity and unpredictability of anomalies make it difficult to collect sufficient training data and the subtle, often context-dependent nature of anomalies complicates their identification using generalized models. Despite these challenges, anomaly detection and prediction are crucial, especially in critical applications like industrial monitoring, financial fraud detection and healthcare, where early and accurate detection of anomalies can prevent significant losses and ensure safety. As time series foundation models continue to evolve, addressing these anomaly detection and prediction challenges will be essential for advancing the field.

\begin{figure*}[!htb]
  \centering
  \includegraphics[width=0.8\linewidth]{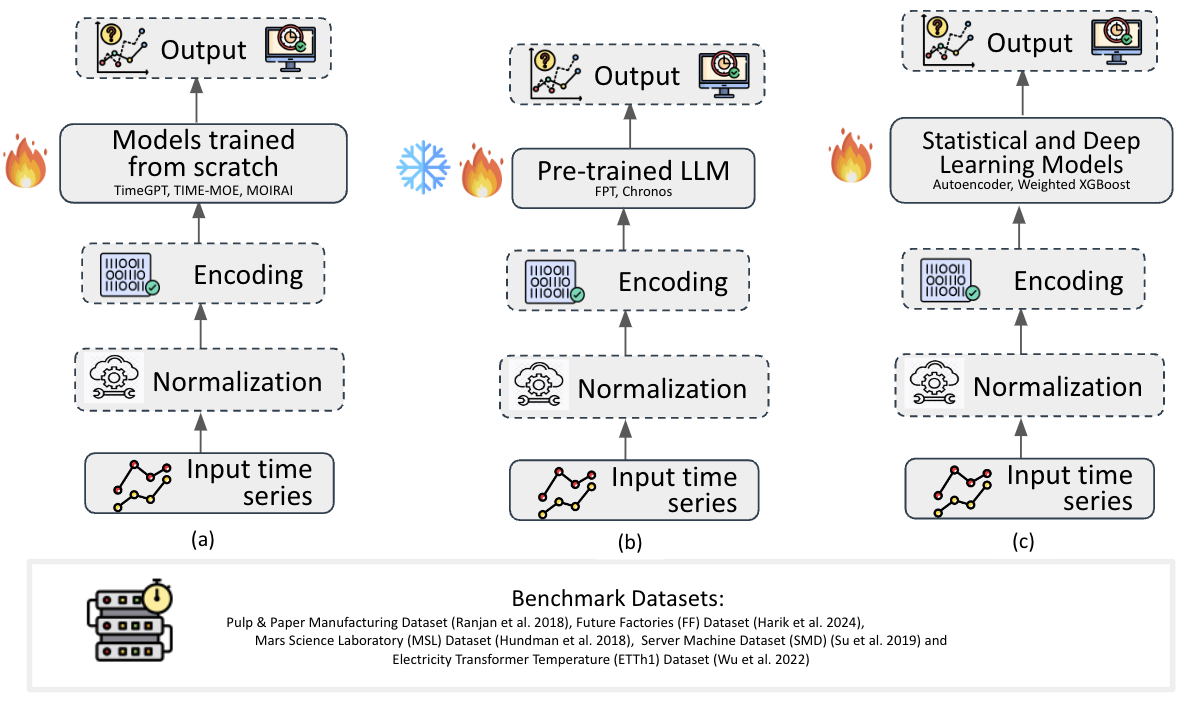}
  \vspace{-2 mm}
\caption{Overview of the analysis procedure. The analysis is categorized into three parts: (a) Foundational models pre-trained from scratch specifically for time series analysis, including TimeGPT \cite{garza2023timegpt}, Time-MOE \cite{shi2024time}, and MOIRAI \cite{gerald2024unified}, all of which can be fine-tuned for additional datasets; (b) Models that adapt large language models (LLMs) for time series tasks, namely FPT \cite{zhou2023one} and Chronos \cite{ansari2024chronos}, where the LLM components remain frozen, with Chronos supporting fine-tuning for other datasets, unlike FPT; and (c) Baseline models trained from scratch for anomaly detection and prediction tasks. The computational cost of all foundational, statistical, and deep learning models used is also evaluated.}
\vspace{-6mm}
  \label{fig:analysis}
\end{figure*}

\vspace{-2mm}
\section{Experimental Setup}
In this section, we outline the setup employed in our experiments, detailing the selection criteria for TSFM, the specific models chosen, the datasets used and the overall process followed to assess the models' performance in time series analysis, particularly in anomaly detection and prediction. 

\vspace{-2mm}
\subsection{Selection of Reference TSFMs}
We chose models for our study based on the criterion that they must be explicitly designed and trained for anomaly detection or forecasting tasks based on multivariate datasets\footnote{For anomaly prediction, we adapt the standard approach of next-time series forecasting.}. Ultimately, we selected three models that involve pre-training foundational models from scratch for time series analysis (TimeGPT\cite{garza2023timegpt}, MOIRAI and Time-MOE\cite{shi2024time}) and two models (Frozen-Pretrained Transformer-FPT\cite{zhou2023one} and Chronos \cite{ansari2024chronos}) that adapts LLMs for time series analysis. Table 1 presents the overview of TSFMs used in this study.

\subsubsection{TimeGPT}
It \cite{garza2023timegpt} utilizes an encoder-decoder transformer architecture. It operates in a univariate channel setting and is designed for detection and forecasting tasks \footnote{https://docs.nixtla.io/}. As the largest transformer-based foundation model in this domain, TimeGPT has been pre-trained on over 100 billion data points. Although it claims to have assembled the largest time series repository from public sources, TimeGPT does not publicly disclose its repository or the details of the data used.

\subsubsection{FPT}
FPT \cite{zhou2023one} utilizes GPT-2 as the backbone. It is a domain-agnostic model primarily focused on forecasting tasks. This approach involves embedding visible LLM adaptation, where most of the LLM parameters are frozen and only a minority are updated during the time series forecasting process. By freezing the major parameters of GPT-2, particularly the self-attention and feedforward layers, FPT preserves a substantial portion of the pre-trained knowledge. The model redesigns the input layer and retrains it, along with the positional embedding and normalization layers, on diverse time series datasets to enhance the LLM's capacity for downstream tasks.

\subsubsection{Time-MOE}
It \cite{shi2024time} introduces a scalable, efficient, and unified architecture for time series forecasting using a Mixture of Experts (MoE) framework. Time-MOE can address challenges in large-scale time series modeling, such as high computational costs and limited flexibility in forecasting horizons. It employs a sparse activation design that reduces computational overhead by activating only a subset of networks for each prediction while maintaining high model capacity. The model uses a decoder-only transformer structure with innovations like multi-resolution forecasting heads, rotary positional embeddings, and Huber loss for stability. Pre-trained on a new large-scale dataset, Time-300B, spanning 300 billion data points across nine domains, Time-MOE demonstrates superior performance in both zero-shot and fine-tuned forecasting tasks. Time-MOE validates the scalability of sparse models, achieving better forecasting accuracy compared to dense counterparts with the same computational resources.

\subsubsection{MOIRAI}
It \cite{gerald2024unified} employs a decoder-only transformer architecture, leveraging a dynamic patching mechanism to tokenize time series data efficiently. Designed for large-scale time series forecasting, MOIRAI integrates a Mixture of Experts framework, enabling it to automatically adapt to varying domain-specific characteristics with minimal tuning. The model uses innovative features such as adaptive patch sizes, which tokenize the input dynamically, and domain-specific mixture-of-experts layers that enhance its generalizability across diverse datasets. Pre-trained on a multi-domain dataset, LOTSA, spanning 27 billion data points, MOIRAI is optimized for both zero-shot and fine-tuned forecasting tasks. Its architecture incorporates automated feature selection and probabilistic forecasting capabilities, allowing it to generate stable predictions even in the presence of high data variability.

\vspace{-2mm}
\subsubsection{Chronos} \cite{ansari2024chronos} Is a framework for pretraining probabilistic models on time-series data by tokenizing these values for use with transformer-based models such as T5. It achieves this through the scaling and quantization of the continuous values of time series into a fixed vocabulary (tokens). Chronos is trained on both publicly available datasets and synthetically generated datasets, which are created using data augmentation techniques such as weighted combinations of publicly available time series data and Gaussian Process Kernels. With models ranging from 20M to 710M parameters, Chronos outperforms statistical and deep learning methods on seen datasets and demonstrates competitive performance on zero-shot datasets.

\vspace{-2mm}
\subsection{Benchmark Datasets}
In this study, we use five publicly available datasets (Appendix A). Four of the datasets are from the manufacturing domain: Pulp and Paper manufacturing dataset \cite{ranjan2018dataset}, Future Factories (FF) dataset ~\cite{harik2024analog}, MSL dataset \cite{hundman2018detecting} and SMD dataset \cite{su2019robust}. These datasets are event-related and include rare events and anomalies. We then use the ETTh1 dataset \cite{wu2022timesnet} \footnote{https://github.com/thuml/Time-Series-Library.git}, which is not event-related.


\vspace{-2mm}
\subsection{Analysis Procedure}
Figure 1 presents the overview of the analysis procedure followed. While experimenting with each model, we replicate the conditions of their original papers. We use the original hyper-parameters, runtime environments, and code, including model architectures, training loops, and data loaders. To ensure a fair comparison, we have included error metrics from the original papers.

\subsubsection{Analysis of Zero-Shot Performance on Anomaly Detection and Prediction}
Firstly, exploratory data analysis (Details in Appendix B) is conducted on each dataset to gain insights into its structure, distribution and key characteristics. Then, data processing techniques (Details in Appendix C) are applied separately to each dataset, tailored to meet the specific requirements of each task. In TimeGPT model, for anomaly detection task, we use NixtlaClient's \textit{"detect\_anomalies"} function to identify anomalies. Additionally, we extend this to a classification task, mapping predicted values to ground truth values included in the dataset for comparison and classifying predictions as normal or abnormal. In anomaly prediction in TimeGPT, we train the model to forecast future data points based on historical data using the \textit{"forecast"} function. In the FPT model, we modify the implementation under \textit{anomaly detection} to perform the detection. For anomaly prediction, we adapt the code in the \textit{long-term forecasting}. For Time-MOE, MOIRAI and Chronos models, we use their zero-shot forecasting capability for anomaly prediction. 

\subsubsection{Analysis of Fine-tuning Performance on Anomaly Detection and Prediction}
We fine-tuned the three reference models, TimeGPT, MOIRAI, and Chronos, on the five datasets employed in our study.

\subsubsection{Analysis with Statistical and Deep Learning Models}
We conduct experiments with statistical and deep learning models as baselines for anomaly detection and prediction. We train a weighted XGBoost model as a statistical method and an autoencoder model that follows encoder-decoder architecture as a deep-learning model. 

\subsubsection{Analysis of Computational Cost of FMs with Statistical and Deep Learning models}
We measure the computational intensity in terms of the time taken to give outputs for the anomaly detection and prediction tasks.

\vspace{-2mm}
\section{Results} 
\subsubsection{Evaluation of Zero-Shot Performance of TSFM in Anomaly Detection and Prediction}
Table 2 presents the evaluation of models across multiple datasets for anomaly detection and prediction tasks \footnote{In the anomaly prediction task, the performance metrics are averaged across 96, 192, 336, and 720 prediction lengths.}. Among the models, only TimeGPT and FPT demonstrate anomaly detection capabilities, with TimeGPT achieving good performance on structured datasets like Pulp (precision: 0.91, recall: 1.0) and MSL datasets (precision: 0.78, F1-score: 0.8). However, its effectiveness diminishes on the FF dataset, where recall falls to 0.06, and prediction metrics such as MSE becomes 0.11. FPT, while less consistent in detection, exhibits reliable forecasting accuracy across datasets like ETTh1 (RMSE: 0.67) and MSL (F1-score: 0.81). Chronos, though limited to forecasting, provides competitive RMSE values on datasets like FF (0.163) and Pulp (0.37), indicating robust point forecast capabilities. For anomaly prediction tasks, MOIRAI demonstrates adaptability but is constrained by its design as a pure forecasting model. On structured datasets like ETTh1 and Pulp, it achieves RMSE values of 0.4 and 0.15, respectively, but struggles with sparse anomaly representations. Time-MOE consistently outperforms other models in forecasting accuracy (e.g., RMSE: 0.587 on ETTh1 and 0.795 on FF) due to its sparse mixture-of-experts framework. Both MOIRAI and Time-MOE highlight the trade-offs between precision and model efficiency in large-scale predictions. Overall, while TimeGPT excels in combined anomaly detection and forecasting, task-specific models like Time-MOE and Chronos deliver competitive results in pure forecasting tasks, reinforcing the need for tailored approaches depending on the dataset characteristics and task requirements.

\begin{table*}[!ht]
\centering
\scriptsize
\renewcommand{\arraystretch}{0.75} 
\label{tab:results_transposed}
\begin{tblr}{
 rowsep = 0pt, 
  row{1} = {c},
  row{2} = {c},
  row{3} = {c},
  row{8} = {c},
  row{13} = {c},
  row{18} = {c},
  row{23} = {c},
  cell{1}{7} = {c=4}{},
  cell{1}{11} = {c=3}{},
  cell{4}{2} = {c},
  cell{4}{3} = {c},
  cell{4}{4} = {c},
  cell{4}{5} = {c},
  cell{4}{6} = {c},
  cell{4}{7} = {c},
  cell{4}{8} = {c},
  cell{4}{9} = {c},
  cell{4}{10} = {c},
  cell{4}{11} = {c},
  cell{4}{12} = {c},
  cell{4}{13} = {c},
  cell{5}{2} = {c},
  cell{5}{3} = {c},
  cell{5}{4} = {c},
  cell{5}{5} = {c},
  cell{5}{6} = {c},
  cell{5}{7} = {c},
  cell{5}{8} = {c},
  cell{5}{9} = {c},
  cell{5}{10} = {c},
  cell{5}{11} = {c},
  cell{5}{12} = {c},
  cell{5}{13} = {c},
  cell{6}{2} = {c},
  cell{6}{3} = {c},
  cell{6}{4} = {c},
  cell{6}{5} = {c},
  cell{6}{6} = {c},
  cell{6}{7} = {c},
  cell{6}{8} = {c},
  cell{6}{9} = {c},
  cell{6}{10} = {c},
  cell{6}{11} = {c},
  cell{6}{12} = {c},
  cell{6}{13} = {c},
  cell{7}{2} = {c},
  cell{7}{3} = {c},
  cell{7}{4} = {c},
  cell{7}{5} = {c},
  cell{7}{6} = {c},
  cell{7}{7} = {c},
  cell{7}{8} = {c},
  cell{7}{9} = {c},
  cell{7}{10} = {c},
  cell{7}{11} = {c},
  cell{7}{12} = {c},
  cell{7}{13} = {c},
  cell{9}{2} = {c},
  cell{9}{3} = {c},
  cell{9}{4} = {c},
  cell{9}{5} = {c},
  cell{9}{6} = {c},
  cell{9}{7} = {c},
  cell{9}{8} = {c},
  cell{9}{9} = {c},
  cell{9}{10} = {c},
  cell{9}{11} = {c},
  cell{9}{12} = {c},
  cell{9}{13} = {c},
  cell{10}{2} = {c},
  cell{10}{3} = {c},
  cell{10}{4} = {c},
  cell{10}{5} = {c},
  cell{10}{6} = {c},
  cell{10}{7} = {c},
  cell{10}{8} = {c},
  cell{10}{9} = {c},
  cell{10}{10} = {c},
  cell{10}{11} = {c},
  cell{10}{12} = {c},
  cell{10}{13} = {c},
  cell{11}{2} = {c},
  cell{11}{3} = {c},
  cell{11}{4} = {c},
  cell{11}{5} = {c},
  cell{11}{6} = {c},
  cell{11}{7} = {c},
  cell{11}{8} = {c},
  cell{11}{9} = {c},
  cell{11}{10} = {c},
  cell{11}{11} = {c},
  cell{11}{12} = {c},
  cell{11}{13} = {c},
  cell{12}{2} = {c},
  cell{12}{3} = {c},
  cell{12}{4} = {c},
  cell{12}{5} = {c},
  cell{12}{6} = {c},
  cell{12}{7} = {c},
  cell{12}{8} = {c},
  cell{12}{9} = {c},
  cell{12}{10} = {c},
  cell{12}{11} = {c},
  cell{12}{12} = {c},
  cell{12}{13} = {c},
  cell{14}{2} = {c},
  cell{14}{3} = {c},
  cell{14}{4} = {c},
  cell{14}{5} = {c},
  cell{14}{6} = {c},
  cell{14}{7} = {c},
  cell{14}{8} = {c},
  cell{14}{9} = {c},
  cell{14}{10} = {c},
  cell{14}{11} = {c},
  cell{14}{12} = {c},
  cell{14}{13} = {c},
  cell{15}{2} = {c},
  cell{15}{3} = {c},
  cell{15}{4} = {c},
  cell{15}{5} = {c},
  cell{15}{6} = {c},
  cell{15}{7} = {c},
  cell{15}{8} = {c},
  cell{15}{9} = {c},
  cell{15}{10} = {c},
  cell{15}{11} = {c},
  cell{15}{12} = {c},
  cell{15}{13} = {c},
  cell{16}{2} = {c},
  cell{16}{3} = {c},
  cell{16}{4} = {c},
  cell{16}{5} = {c},
  cell{16}{6} = {c},
  cell{16}{7} = {c},
  cell{16}{8} = {c},
  cell{16}{9} = {c},
  cell{16}{10} = {c},
  cell{16}{11} = {c},
  cell{16}{12} = {c},
  cell{16}{13} = {c},
  cell{17}{2} = {c},
  cell{17}{3} = {c},
  cell{17}{4} = {c},
  cell{17}{5} = {c},
  cell{17}{6} = {c},
  cell{17}{7} = {c},
  cell{17}{8} = {c},
  cell{17}{9} = {c},
  cell{17}{10} = {c},
  cell{17}{11} = {c},
  cell{17}{12} = {c},
  cell{17}{13} = {c},
  cell{19}{2} = {c},
  cell{19}{3} = {c},
  cell{19}{4} = {c},
  cell{19}{5} = {c},
  cell{19}{6} = {c},
  cell{19}{7} = {c},
  cell{19}{8} = {c},
  cell{19}{9} = {c},
  cell{19}{10} = {c},
  cell{19}{11} = {c},
  cell{19}{12} = {c},
  cell{19}{13} = {c},
  cell{20}{2} = {c},
  cell{20}{3} = {c},
  cell{20}{4} = {c},
  cell{20}{5} = {c},
  cell{20}{6} = {c},
  cell{20}{7} = {c},
  cell{20}{8} = {c},
  cell{20}{9} = {c},
  cell{20}{10} = {c},
  cell{20}{11} = {c},
  cell{20}{12} = {c},
  cell{20}{13} = {c},
  cell{21}{2} = {c},
  cell{21}{3} = {c},
  cell{21}{4} = {c},
  cell{21}{5} = {c},
  cell{21}{6} = {c},
  cell{21}{7} = {c},
  cell{21}{8} = {c},
  cell{21}{9} = {c},
  cell{21}{10} = {c},
  cell{21}{11} = {c},
  cell{21}{12} = {c},
  cell{21}{13} = {c},
  cell{22}{2} = {c},
  cell{22}{3} = {c},
  cell{22}{4} = {c},
  cell{22}{5} = {c},
  cell{22}{6} = {c},
  cell{22}{7} = {c},
  cell{22}{8} = {c},
  cell{22}{9} = {c},
  cell{22}{10} = {c},
  cell{22}{11} = {c},
  cell{22}{12} = {c},
  cell{22}{13} = {c},
  cell{24}{2} = {c},
  cell{24}{3} = {c},
  cell{24}{4} = {c},
  cell{24}{5} = {c},
  cell{24}{6} = {c},
  cell{24}{7} = {c},
  cell{24}{8} = {c},
  cell{24}{9} = {c},
  cell{24}{10} = {c},
  cell{24}{11} = {c},
  cell{24}{12} = {c},
  cell{24}{13} = {c},
  cell{25}{2} = {c},
  cell{25}{3} = {c},
  cell{25}{4} = {c},
  cell{25}{5} = {c},
  cell{25}{6} = {c},
  cell{25}{7} = {c},
  cell{25}{8} = {c},
  cell{25}{9} = {c},
  cell{25}{10} = {c},
  cell{25}{11} = {c},
  cell{25}{12} = {c},
  cell{25}{13} = {c},
  cell{26}{2} = {c},
  cell{26}{3} = {c},
  cell{26}{4} = {c},
  cell{26}{5} = {c},
  cell{26}{6} = {c},
  cell{26}{7} = {c},
  cell{26}{8} = {c},
  cell{26}{9} = {c},
  cell{26}{10} = {c},
  cell{26}{11} = {c},
  cell{26}{12} = {c},
  cell{26}{13} = {c},
  cell{27}{2} = {c},
  cell{27}{3} = {c},
  cell{27}{4} = {c},
  cell{27}{5} = {c},
  cell{27}{6} = {c},
  cell{27}{7} = {c},
  cell{27}{8} = {c},
  cell{27}{9} = {c},
  cell{27}{10} = {c},
  cell{27}{11} = {c},
  cell{27}{12} = {c},
  cell{27}{13} = {c},
  vline{2-8} = {1}{},
  vline{2-7,11} = {2-27}{},
  hline{1-3,8,13,18,23,28} = {-}{},
}
        &            &                &         &       &        & Anomaly Detection &        &          &          & Anomaly Prediction &       &       \\
Dataset & \#Features & TSFM           & \#Train & \#Val & \#Test & Precision         & Recall & F1-Score & Accuracy & MSE                & RMSE  & MAE   \\
FF      & 3          & TimeGPT        & 779K    &       & 194K   & 0.84              & 0.06   & 0.11     & 0.56     & 0.11               & 0.32  & 0.25  \\
        & 20         & FPT     & 682K    & 195K  & 97K    & 0.09              & 0.84   & 0.16     & 0.98     & 0.27               & 0.53  & 0.17  \\
        & 20         & Time-MOE (base)       & 779K    &       & 194K   & -                 & -      & -        & -        & 0.6325             & 0.795 & 0.349 \\
        & 20         & MOIRAI (base)  & 779K    &       & 194K   & -                 & -      & -        & -        & 0.06               & 0.244 & 0.05  \\
        & 20         & Chronos (tiny) & 779K    &       & 194K   & -                 & -      & -        & -        & 0.027              & 0.163 & 0.097 \\
Pulp    & 59         & TimeGPT        & 14718   &       & 3680   & 0.91              & 1      & 0.95     & 0.99     & 0.20               & 0.45  & 0.37  \\
        & 59         & FPT     & 12878   & 3680  & 1840   & 0.01              & 0.04   & 0.02     & 0.97     & 0.65               & 0.806 & 1.77  \\
        & 20         & Time-MOE (base)       & 14718   &       & 3680   & -                 & -      & -        & -        & 2.2043             & 1.484 & 0.406 \\
        & 20         & MOIRAI (base)  & 14718   &       & 3680   & -                 & -      & -        & -        & 0.02               & 0.15  & 0.041 \\
        & 20         & Chronos (tiny) &         &       &        & -                 & -      & -        & -        & 0.13               & 0.37  & 0.25  \\
SMD     & 37         & TimeGPT        & 60701   &       & 15175  & 0.7               & 0.74   & 0.81     & 0.78     & 0.20               & 0.45  & 0.40  \\
        & 37         & FPT     & 45526   & 15175 & 15175  & 0.82              & 0.81   & 0.8      & 0.9      & 0.98               & 0.98  & 2.77  \\
        & 37         & Time-MOE (base)      & 60701   &       & 15175  & -                 & -      & -        & -        & 1.3                & 1.484 & 0.6   \\
        & 37         & MOIRAI (base)  & 60701   &       & 15175  & -                 & -      & -        & -        & 0.011              & 0.106 & 0.017 \\
        & 37         & Chronos (tiny) & 60701   &       & 15175  & -                 & -      & -        & -        & 0.15               & 0.39  & 0.25  \\
MSL     & 55         & TimeGPT        & 56K     &       & 73K    & 0.78              & 0.77   & 0.8      & 0.78     & 0.5                & 0.70  & 0.52  \\
        & 55         & FPT     & 56K     &       & 73K    & 0.81              & 0.81   & 0.8142   & 0.96     & 0.2                & 0.51  & 0.19  \\
        & 55         & Time-MOE (base)       & 56K     &       & 73K    & -                 & -      & -        & -        & 0.04               & 0.2   & 0.09  \\
        & 55         & MOIRAI (base)  & 56K     &       & 73K    & -                 & -      & -        & -        & 1.2                & 1.09  & 2.3   \\
        & 55         & Chronos (tiny) & 56K     &       & 73K    & -                 & -      & -        & -        & 0.04               & 0.19  & 0.14  \\
ETTh1   & 7          & TimeGPT        & 13937   &       & 3484   & -                 & -      & -        & -        & 0.24               & 0.49  & 0.46  \\
        & 7          & FPT     & 12195   & 3484  & 1742   & -                 & -      & -        & -        & 0.46               & 0.67  & 0.47  \\
        & 7          & Time-MOE (base)      & 13937   &       & 3484   & -                 & -      & -        & -        & 0.345              & 0.587 & 0.373 \\
        & 7          & MOIRAI (base)  & 14718   &       & 3680   & -                 & -      & -        & -        & 0.2                & 0.44  & 0.41  \\
        & 7          & Chronos (tiny) & 14718   &       & 3680   & -                 & -      & -        & -        & 0.01               & 0.11  & 0.02  
\end{tblr}
\caption{Experimental Results of Dataset-Wise Model Analysis for Zero-Shot Anomaly Detection and Prediction.
FF: Future factories dataset, Pulp: Pulp and paper manufacturing dataset, SMD: Server machine dataset, MSL: Mars science laboratory dataset,  ETTh1: Electricity transformer temperature dataset,
TSFM: Time series foundational model, MSE: Mean squared error, RMSE: Root mean squared error, MAE: Mean absolute error}
\vspace{-6mm}
\end{table*}

\begin{table*}
\scriptsize
\renewcommand{\arraystretch}{0.75} 
\centering
\label{tab:results_transposed}
\begin{tblr}{
rowsep = 0.5pt, 
  row{1} = {c},
  row{2} = {c},
  row{3} = {c},
  row{6} = {c},
  row{9} = {c},
  row{12} = {c},
  row{15} = {c},
  cell{1}{6} = {c=4}{},
  cell{1}{10} = {c=3}{},
  cell{4}{2} = {c},
  cell{4}{3} = {c},
  cell{4}{4} = {c},
  cell{4}{5} = {c},
  cell{4}{6} = {c},
  cell{4}{7} = {c},
  cell{4}{8} = {c},
  cell{4}{9} = {c},
  cell{4}{10} = {c},
  cell{4}{11} = {c},
  cell{4}{12} = {c},
  cell{5}{2} = {c},
  cell{5}{3} = {c},
  cell{5}{4} = {c},
  cell{5}{5} = {c},
  cell{5}{6} = {c},
  cell{5}{7} = {c},
  cell{5}{8} = {c},
  cell{5}{9} = {c},
  cell{5}{10} = {c},
  cell{5}{11} = {c},
  cell{5}{12} = {c},
  cell{7}{2} = {c},
  cell{7}{3} = {c},
  cell{7}{4} = {c},
  cell{7}{5} = {c},
  cell{7}{6} = {c},
  cell{7}{7} = {c},
  cell{7}{8} = {c},
  cell{7}{9} = {c},
  cell{7}{10} = {c},
  cell{7}{11} = {c},
  cell{7}{12} = {c},
  cell{8}{2} = {c},
  cell{8}{3} = {c},
  cell{8}{4} = {c},
  cell{8}{5} = {c},
  cell{8}{6} = {c},
  cell{8}{7} = {c},
  cell{8}{8} = {c},
  cell{8}{9} = {c},
  cell{8}{10} = {c},
  cell{8}{11} = {c},
  cell{8}{12} = {c},
  cell{10}{2} = {c},
  cell{10}{3} = {c},
  cell{10}{4} = {c},
  cell{10}{5} = {c},
  cell{10}{6} = {c},
  cell{10}{7} = {c},
  cell{10}{8} = {c},
  cell{10}{9} = {c},
  cell{10}{10} = {c},
  cell{10}{11} = {c},
  cell{10}{12} = {c},
  cell{11}{2} = {c},
  cell{11}{3} = {c},
  cell{11}{4} = {c},
  cell{11}{5} = {c},
  cell{11}{6} = {c},
  cell{11}{7} = {c},
  cell{11}{8} = {c},
  cell{11}{9} = {c},
  cell{11}{10} = {c},
  cell{11}{11} = {c},
  cell{11}{12} = {c},
  cell{13}{2} = {c},
  cell{13}{3} = {c},
  cell{13}{4} = {c},
  cell{13}{5} = {c},
  cell{13}{6} = {c},
  cell{13}{7} = {c},
  cell{13}{8} = {c},
  cell{13}{9} = {c},
  cell{13}{10} = {c},
  cell{13}{11} = {c},
  cell{13}{12} = {c},
  cell{14}{2} = {c},
  cell{14}{3} = {c},
  cell{14}{4} = {c},
  cell{14}{5} = {c},
  cell{14}{6} = {c},
  cell{14}{7} = {c},
  cell{14}{8} = {c},
  cell{14}{9} = {c},
  cell{14}{10} = {c},
  cell{14}{11} = {c},
  cell{14}{12} = {c},
  cell{16}{2} = {c},
  cell{16}{3} = {c},
  cell{16}{4} = {c},
  cell{16}{5} = {c},
  cell{16}{6} = {c},
  cell{16}{7} = {c},
  cell{16}{8} = {c},
  cell{16}{9} = {c},
  cell{16}{10} = {c},
  cell{16}{11} = {c},
  cell{16}{12} = {c},
  cell{17}{2} = {c},
  cell{17}{3} = {c},
  cell{17}{4} = {c},
  cell{17}{5} = {c},
  cell{17}{6} = {c},
  cell{17}{7} = {c},
  cell{17}{8} = {c},
  cell{17}{9} = {c},
  cell{17}{10} = {c},
  cell{17}{11} = {c},
  cell{17}{12} = {c},
  vline{2-7} = {1}{},
  vline{2-6,10} = {2-17}{},
  hline{1-3,6,9,12,15,18} = {-}{},
}
        &            &                &         &        & Anomaly Detection &        &          &          & Anomaly Prediction &       &       \\
Dataset & \#Features & TSFM           & \#Train & \#Test & Precision         & Recall & F1-Score & Accuracy & MSE                & RMSE  & MAE   \\
FF      & 3          & TimeGPT        & 779K    & 194K   & 0.85              & 0.08   & 0.13     & 0.58     & 0.10               & 0.31  & 0.25  \\
        & 20         & MOIRAI (base)  & 779K    & 194K   & -                 & -      & -        & -        & 0.05               & 0.22  & 0.05  \\
        & 20         & Chronos (tiny) & 779K    & 194K   & -                 & -      & -        & -        & 0.026              & 0.16  & 0.098 \\
Pulp    & 59         & TimeGPT        & 14718   & 3680   & 0.92              & 1      & 0.95     & 0.99     & 0.18               & 0.42  & 0.34  \\
        & 20         & MOIRAI (base)  & 14718   & 3680   & -                 & -      & -        & -        & 0.015              & 0.122 & 0.04  \\
        & 20         & Chronos (tiny) & 14718   & 3680   & -                 & -      & -        & -        & 0.13               & 0.36  & 0.23  \\
SMD     & 37         & TimeGPT        & 60701   & 15175  & -                 & -      & -        & -        & 0.14               & 0.38  & 0.32  \\
        & 37         & MOIRAI (base)  & 60701   & 15175  & -                 & -      & -        & -        & 1.3                & 1.484 & 0.6   \\
        & 37         & Chronos (tiny) & 60701   & 15175  & -                 & -      & -        & -        & 0.13               & 0.36  & 0.23  \\
MSL     & 55         & TimeGPT        & 56K     & 73K    & 0.79              & 0.77   & 0.8      & 0.78     & 0.49               & 0.7   & 0.52  \\
        & 55         & MOIRAI (base)  & 56K     & 73K    & -                 & -      & -        & -        & 1.1                & 1.04  & 2.1   \\
        & 55         & Chronos (tiny) & 56K     & 73K    & -                 & -      & -        & -        & 0.03               & 0.12  & 0.02  \\
ETTh1   & 7          & TimeGPT        & 14718   & 3680   & -                 & -      & -        & -        & 0.24               & 0.49  & 0.45  \\
        & 7          & MOIRAI (base)  & 14718   & 3680   & -                 & -      & -        & -        & 0.2                & 0.44  & 0.41  \\
        & 7          & Chronos (tiny) & 14718   & 3680   & -                 & -      & -        & -        & 0.01               & 0.11  & 0.02  
\end{tblr}
\caption{Experimental Results of Dataset-Wise Model Analysis of Fine-tuning for Anomaly Detection and Prediction.
FF: Future factories dataset, Pulp: Pulp and paper manufacturing dataset, SMD: Server machine dataset, MSL: Mars science laboratory dataset,  ETTh1: Electricity transformer temperature dataset,
TSFM: Time series foundational model, MSE: Mean squared error, RMSE: Root mean squared error, MAE: Mean absolute error}
\vspace{-4mm}
\end{table*}

\begin{table*}[!ht]
\scriptsize
\renewcommand{\arraystretch}{0.75} 
\centering
\label{tab:results_transposed}
\begin{tblr}{
  cells = {c},
  rowsep = 0.5pt,
  cell{1}{6} = {c=4}{},
  cell{1}{10} = {c=3}{},
  vline{2-7} = {1}{},
  vline{2-6,10} = {2-12}{},
  hline{1-3,5,7,9,11,13} = {-}{},
}
        &            &              &         &        & Anomaly Detection &        &          &          & Anomaly Prediction &       &      \\
Dataset & \#Features & Model        & \#Train & \#Test & Precision         & Recall & F1-Score & Accuracy & mse                & rmse  & mae  \\
FF      & 20         & Autoencoder  & 779K    & 194K   & 0.6               & 0.61   & 0.63     & 0.62     & 2.5                & 1.58  & 1.6  \\
        &            & Weighted XGB & 779K    & 194K   & 0.81              & 0.82   & 0.8      & 0.81     & 0.2                & 0.44  & 0.29 \\
Pulp    & 59         & Autoencoder  & 14718   & 3680   & 0.89              & 0.88   & 0.893    & 0.87     & 0.4                & 0.63  & 0.3  \\
        &            & Weighted XGB & 14718   & 3680   & 0.9               & 0.93   & 0.94     & 0.94     & 0.05               & 0.22  & 0.12 \\
SMD     & 37         & Autoencoder  & 60701   & 15175  & 0.79              & 0.78   & 0.79     & 0.78     & 0.3                & 0.54  & 0.2  \\
        &            & Weighted XGB & 60701   & 15175  & 0.85              & 0.86   & 0.87     & 0.88     & 0.08               & 0.28  & 0.93 \\
MSL     & 55         & Autoencoder  & 56K     & 73K    & 0.72              & 0.71   & 0.73     & 0.72     & 1.2                & 1.095 & 1.3  \\
        &            & Weighted XGB & 56K     & 73K    & 0.8               & 0.82   & 0.8      & 0.81     & 0.3                & 0.54  & 0.7  \\
ETTh1   & 7          & Autoencoder  & 13937   & 3484   & -                 & -      & -        & -        & 2.4                & 1.55  & 2.1  \\
        &            & Weighted XGB & 13937   & 3484   & -                 & -      & -        & -        & 0.92               & 0.95  & 1.2  
\end{tblr}
\caption{Experimental Results with Baselines.
FF: Future factories dataset, Pulp: Pulp and paper manufacturing dataset, SMD: Server machine dataset, MSL: Mars science laboratory dataset,  ETTh1: Electricity transformer temperature dataset,
TSFM: Time series foundational model, MSE: Mean squared error, RMSE: Root mean squared error, MAE: Mean absolute error}
\vspace{-6mm}
\end{table*}

\subsubsection{Evaluation of Fine-tuning Performance of TSFM in Anomaly Detection and Prediction}
Fine-tuning of TimeGPT, MOIRAI, and Chronos demonstrated only marginal improvements in performance compared to their zero-shot results, as highlighted in Table 3. This indicates that the benefits of fine-tuning these models for anomaly detection and prediction tasks are limited. 

\subsubsection{Evaluation of the Analysis with Statistical and Deep Learning Models}
Table 4 presents a comparative analysis of statistical(Weighted XGBoost) and deep learning (Autoencoder) models across the datasets we used for anomaly detection and prediction tasks. In terms of classification metrics, Weighted XGB consistently outperforms Autoencoder across datasets, demonstrating superior precision, recall, F1-scores, and accuracy. For example, on the Pulp dataset, Weighted XGB achieves an F1-score of 0.94 compared to Autoencoder's 0.893, with an improvement in recall (0.93 vs. 0.88) and precision (0.9 vs. 0.89). Similarly, on the FF dataset, Weighted XGB achieves higher F1-scores (0.8 vs. 0.63) and accuracy (0.81 vs. 0.62). When assessing regression performance, Weighted XGB also outperforms Autoencoder with lower MSE, RMSE, and MAE values across datasets, indicating better anomaly prediction capabilities. For instance, in the Pulp dataset, Weighted XGB achieves an MSE of 0.05 compared to Autoencoder's 0.4 and an RMSE of 0.22 compared to 0.63. A similar trend is observed in other datasets, such as MSL and SMD, where Weighted XGB consistently yields better predictive accuracy. Notably, the ETTh1 dataset shows Weighted XGB achieving significantly lower RMSE (0.95) and MAE (1.2) values compared to Autoencoder. These results highlight Weighted XGB’s superior efficiency and accuracy in both anomaly detection and prediction tasks across diverse datasets. When compared to TSFMs, the results indicate that statistical and deep learning models consistently outperform time series foundational models in both anomaly detection and prediction tasks.

\vspace{-4mm}
\subsubsection{Evaluation of the Analysis of Computational Cost of FMs, Statistical and Deep Learning Models}
Table 5 (Appendix E) presents the comparative evaluation of the reference TSFMs for zero-shot anomaly detection and prediction tasks across five datasets we used with differing feature dimensions. Across all the datasets, the Autoencoder model and the Weighted XGBoost model demonstrate superior computational efficiency, consistently achieving the lowest anomaly detection and prediction inference times, often within a range of 0.54 to 0.85 minutes. In contrast, more complex and generalized models like FPT exhibit significantly longer processing times, exceeding 30 minutes in most cases, highlighting a trade-off between generalizability and computational cost. TimeGPT, while moderately efficient, strikes a balance by achieving competitive performance in detection and prediction times, ranging from 6.1 to 15.1 minutes and 8.2 to 24.7 minutes, respectively. Time-MOE, MOIRAI, and Chronos models demonstrate faster performance compared to TimeGPT and FPT models in anomaly prediction tasks. However, their processing times remain slower when compared to the highly efficient statistical and deep learning models such as Autoencoder and Weighted XGBoost. This analysis underscores the variability in model performance depending on dataset characteristics, highlighting the suitability of lightweight models like Autoencoder and Weighted XGBoost for time-sensitive anomaly detection tasks, while models like TimeGPT offer a middle ground between computational efficiency and adaptability.

\vspace{-4mm}
\section{Discussion}

In this paper, we examined the performance of foundational models for anomaly detection and prediction. Despite their growing popularity, our findings indicate that foundational models do not substantially enhance performance compared to other methodologies. Our results underscore the performance of TSFMs in anomaly detection and prediction and reveal that statistical and deep learning models consistently outperform them in performance and computational efficiency. Statistical models like weighted XGBoost excel in anomaly detection and prediction due to their interpretability, using feature importance and ensemble methods to identify variables contributing to anomalies. Deep learning methods, such as autoencoders, learn compressed representations of "normal" data, making them detect deviations as anomalies effectively. In contrast, TSFMs designed for general-purpose forecasting often struggle to capture the task-specific nuances required for effective anomaly detection or prediction. Furthermore, only a few models, such as TimeGPT and FPT, are extended for anomaly detection, and none are purpose-built for anomaly prediction, where we have repurposed forecasting capabilities.

To address these challenges, future research on TSFMs must focus on integrating robust mechanisms for anomaly detection and prediction. Novel tokenization, data augmentation, and prompting techniques could be developed to enhance foundational model learning. Scaling laws and domain knowledge integration should be leveraged to capture complex temporal dynamics effectively, optimizing model performance and reducing computational complexity. Expanding into multimodal datasets represents another promising direction, enabling the integration of diverse data modalities. Additionally, improving model explainability is critical. The black-box nature of TSFMs limits their interpretability, an essential factor in high-stakes applications such as healthcare and manufacturing. Techniques such as chain-of-thought prompting or post-hoc explanation methods, which have proven effective in LLMs, remain underutilized in TSFMs.

\vspace{-4mm}
\section{Conclusion}
We show that TSFMs, while effective in forecasting, are limited in anomaly detection and prediction. Statistical and deep learning models outperform TSFMs regarding accuracy, efficiency, and usability. The black-box nature of TSFMs and the lack of specialized designs for anomaly-related tasks restrict their applicability. Future advancements should focus on integrating domain knowledge, multimodal data, and explainability techniques to improve TSFM performance. Addressing these gaps will make TSFMs more practical for critical applications such as industrial monitoring and financial fraud detection.

\bibliography{aaai25}

\clearpage
\onecolumn

\section{Appendix}


\subsection{Appendix A. Description of the Datasets Used}
\label{sec:appendixA1}
\subsubsection{Pulp and paper manufacturing dataset}
This dataset has been curated from the pulp-and-paper industry and contains sensor data from various locations on machines within a paper mill  \cite{ranjan2018dataset}. These sensors monitor metrics of raw materials, such as pulp fiber and chemicals, as well as process variables like blade type, couch vacuum and rotor speed. The dataset comprises 18,398 records over a 30-day period, with sensor readings taken every two minutes. Each record includes a timestamp, a binary event label (y) and 61 predictor variables (x1–x61). Most predictors are continuous, except for x28, which is categorical and x61, which is binary. Only 124 rows have a y value of 1, indicating a sheet break (a rare event), while the rest have a value of 0.

\subsubsection{Future factories dataset}
This is a publicly available manufacturing dataset curated by the Future Factories (FF) lab at the University of South Carolina \cite{harik2024analog}. This dataset is available in two formats: analog and multimodal. It contains data captured from a prototype rocket assembly pipeline, which follows industrial standards for the deployment of actuators, control systems and transducers. In this study, we use the analog FF dataset \footnote{https://github.com/smtmnfg/NSF-MAP}. The dataset encompasses 292 complete assembly cycles, where each cycle represents the full assembly and disassembly of a rocket. The assembly process at the FF lab is segmented into 21 distinct cycle states. The preprocessed dataset includes various measurements, such as sensor readings, conveyor variable frequency drive temperatures, robot physical properties (e.g., angles), conveyor workstation statistics, cycle states, cycle counts, anomaly types and associated image file names from both cameras. The rocket assembled in the FF Lab consists of four parts: the nosecone, body 1, body 2 and the rocket base. Any missing part is categorized as an anomaly; for instance, the absence of Rocket body 1 is labeled "NoBody1," while the absence of both Rocket body 1 and body 2 is labeled "NoBody1,NoBody2."

\subsubsection{MSL dataset}
The MSL dataset \cite{hundman2018detecting} \footnote{https://github.com/khundman/telemanom
https://arxiv.org/pdf/1802.04431} includes 66,709 data records of anonymized spacecraft telemetry channel data and anomalies. Channel IDs are anonymized, but the first letter indicates the type of channel (e.g., P = power, R = radiation). There are total 36 anomalous records in the dataset, divided into two categories: 19 point-anomalies, which ignore temporal information and are distance-based and 17 contextual anomalies, identified by properly-set alarms that take temporal information into consideration. No identifying information related to the timing or nature of commands is included in the data. Additionally, there are 27 unique features associated with telemetry channels collected via manual telemetry assessment and 19 unique features obtained by mining ISA reports.

\subsubsection{Server machine dataset (SMD)}
The SMD\footnote{\url{https://github.com/NetManAIOps/OmniAnomaly}} is a comprehensive five-week dataset collected from a large Internet company, designed to support research on anomaly detection and interpretability in machine learning. It comprises data from 28 individual machines, grouped into three distinct categories, each labeled using the format \texttt{machine-<group\_index>-<index>}. For each machine, the dataset is divided into two equal parts: the first half serves as the training set, while the second half constitutes the test set. To facilitate evaluation, the dataset provides test labels indicating whether a data point is anomalous and interpretation labels that identify the specific dimensions contributing to each anomaly. The structured design of SMD, which emphasizes independent training and testing for each machine, makes it a valuable resource for developing and benchmarking advanced anomaly detection algorithms and interpretability techniques. SMD consists of multivariate time series data with a dimensionality of 38, a training set containing 708,405 data points, and a testing set comprising 708,420 data points. The anomaly ratio within the dataset is approximately 4.16\%. For our analysis, we used a part of the SMD dataset.

\subsubsection{ETTh1 dataset}
The ETTh1 dataset is a subset of the Electricity Transformer Temperature (ETT) dataset \cite{wu2022timesnet}, which focuses on time-series data collected hourly from electricity transformers. This dataset includes two years of data, amounting to 17,520 data points, with each entry consisting of eight features: "date", six external power load features (HUFL, HULL, MUFL, MULL, LUFL, LULL) and the target variable, "oil temperature" (OT). ETTh1 is designed to capture short-term daily and long-term weekly patterns, making it suitable for forecasting tasks in electrical transformer management, such as predicting transformer oil temperature and optimizing load distribution.

\clearpage


\subsection{Appendix B. Exploratory Data Analysis (EDA)}
\label{sec:AppendixA4}
\begin{figure}[!htb]
  \centering
  \begin{subfigure}[b]{0.45\textwidth}
    \includegraphics[width=\textwidth]{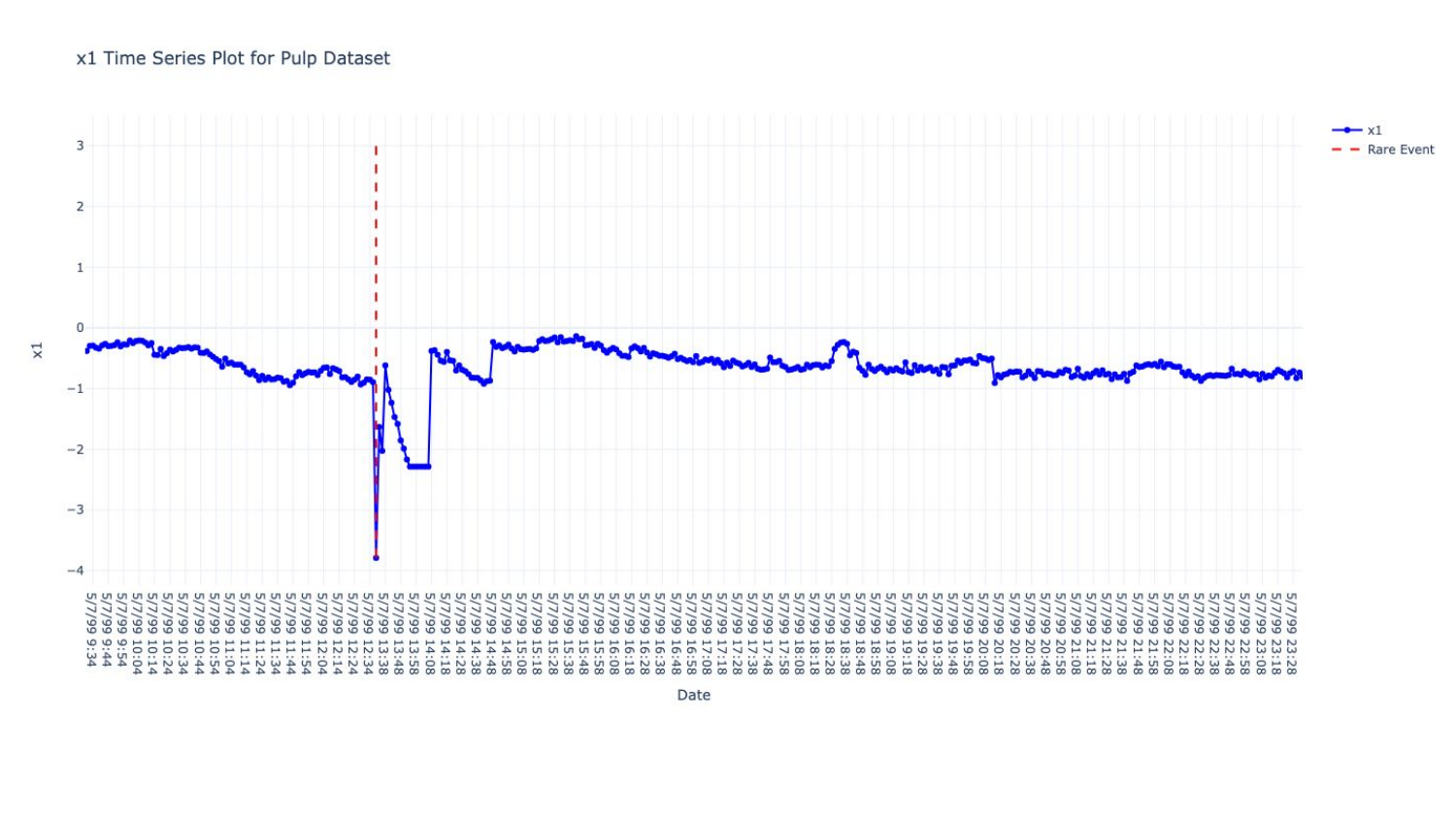}
    \caption{Time Series Plot for Pulp Dataset-Feature x1. The dashed red line presents the occurrence of a rare event.}
    \label{fig:x1}
  \end{subfigure}
  \hfill
  \begin{subfigure}[b]{0.45\textwidth}
    \includegraphics[width=\textwidth]{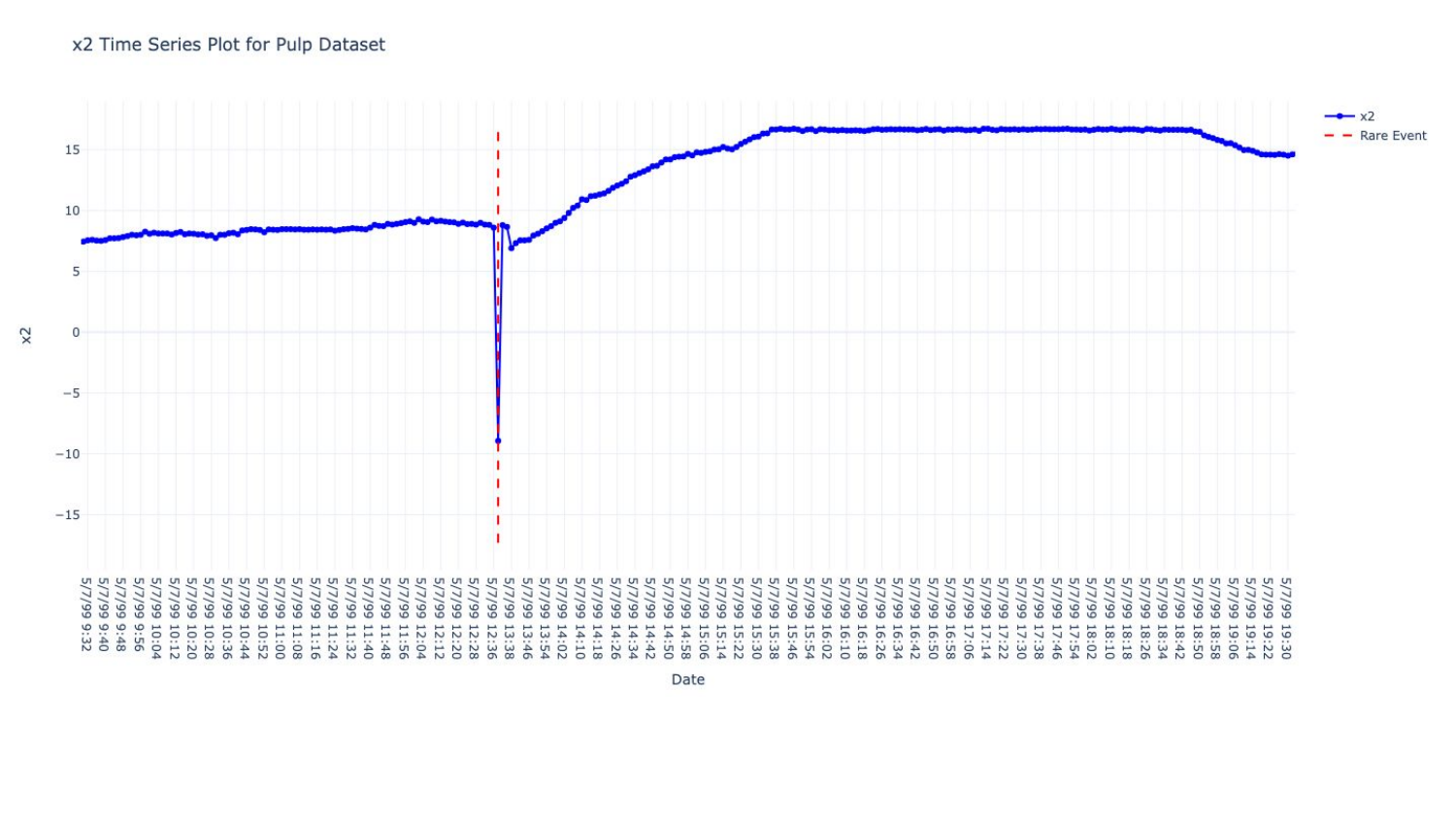}
    \caption{Time Series Plot for Pulp Dataset-Feature x2. The dashed red line presents the occurrence of a rare event.}
    \label{fig:x2}
  \end{subfigure}

  \vspace{0.5cm} 

  \begin{subfigure}[b]{0.45\textwidth}
    \includegraphics[width=\textwidth]{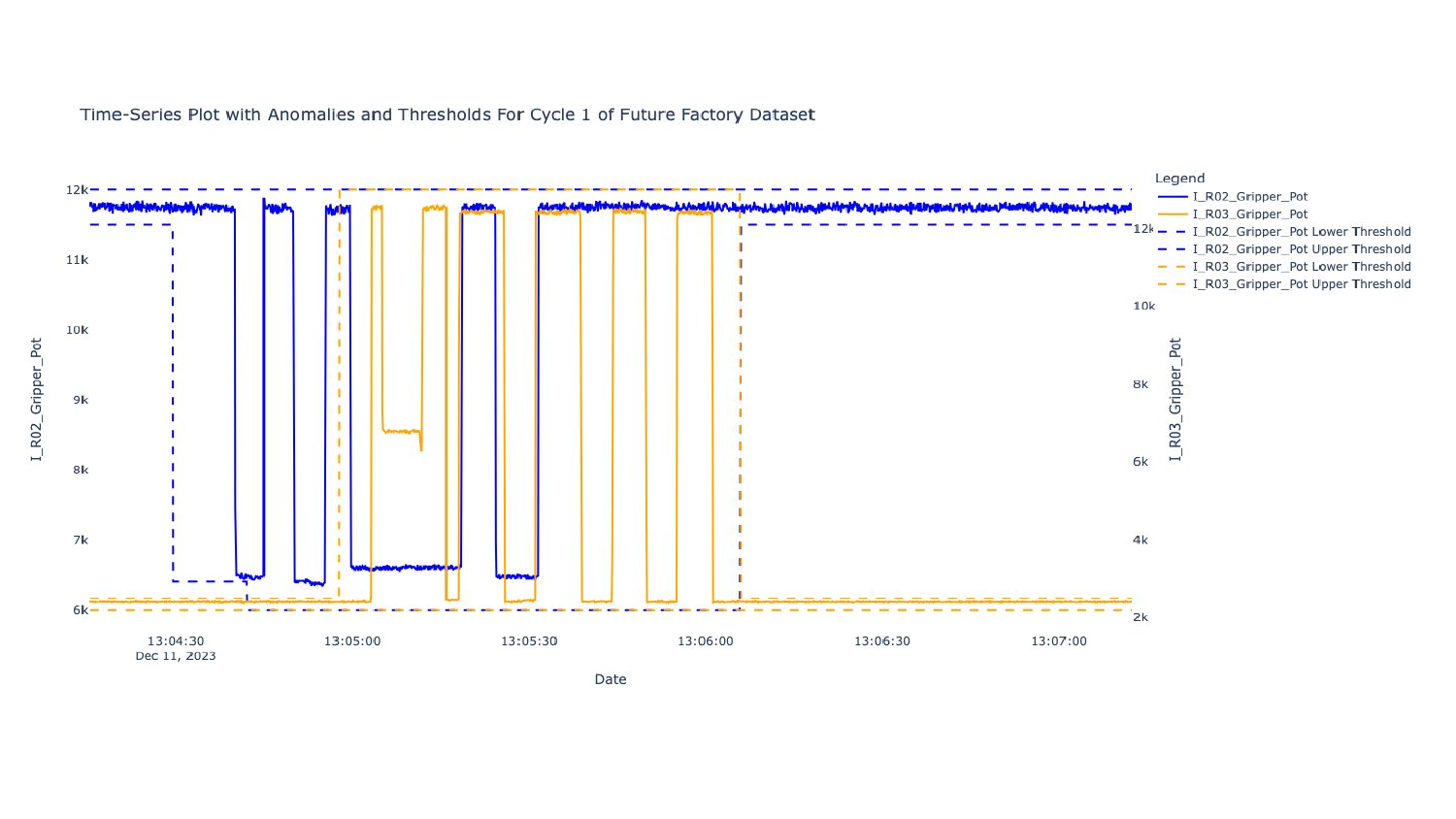}
    \caption{Time series plot for FF dataset with anomalies and thresholds for a selected \textbf{Normal} cell cycle. The dashed horizontal lines represent the upper and lower thresholds of the features.}
    \label{fig:FF5}
  \end{subfigure}
  \hfill
  \begin{subfigure}[b]{0.45\textwidth}
    \includegraphics[width=\textwidth]{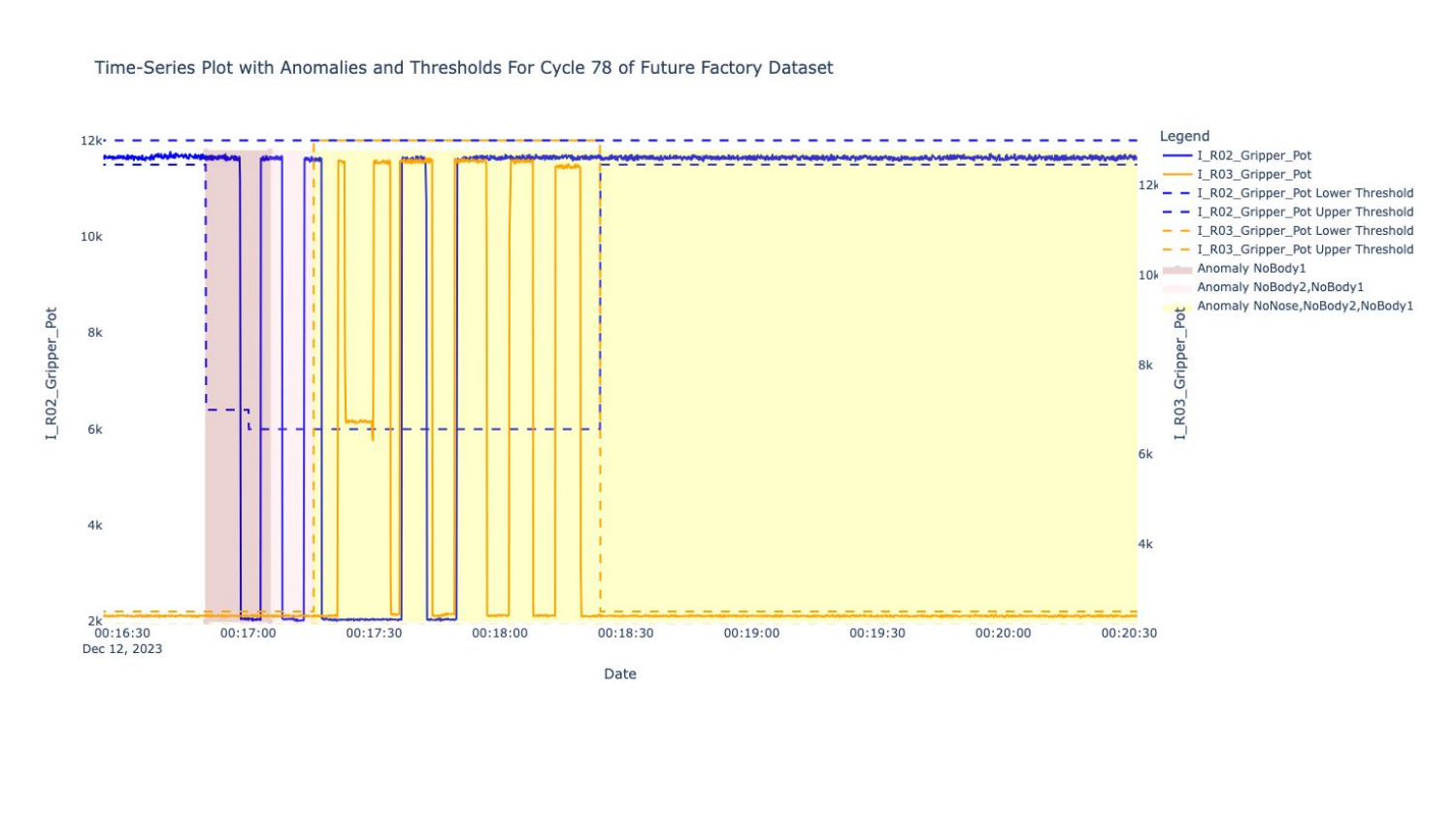}
    \caption{Time series plot for FF dataset with anomalies and thresholds for a selected \textbf{Anomalous} cell cycle. The dashed horizontal lines represent the upper and lower thresholds of the features. The vertical colored segments indicate the presence of anomalies, with each color representing a different anomaly type.}
    \label{fig:FF6}
  \end{subfigure}

  \vspace{0.5cm} 

  \begin{subfigure}[b]{0.45\textwidth}
    \includegraphics[width=\textwidth]{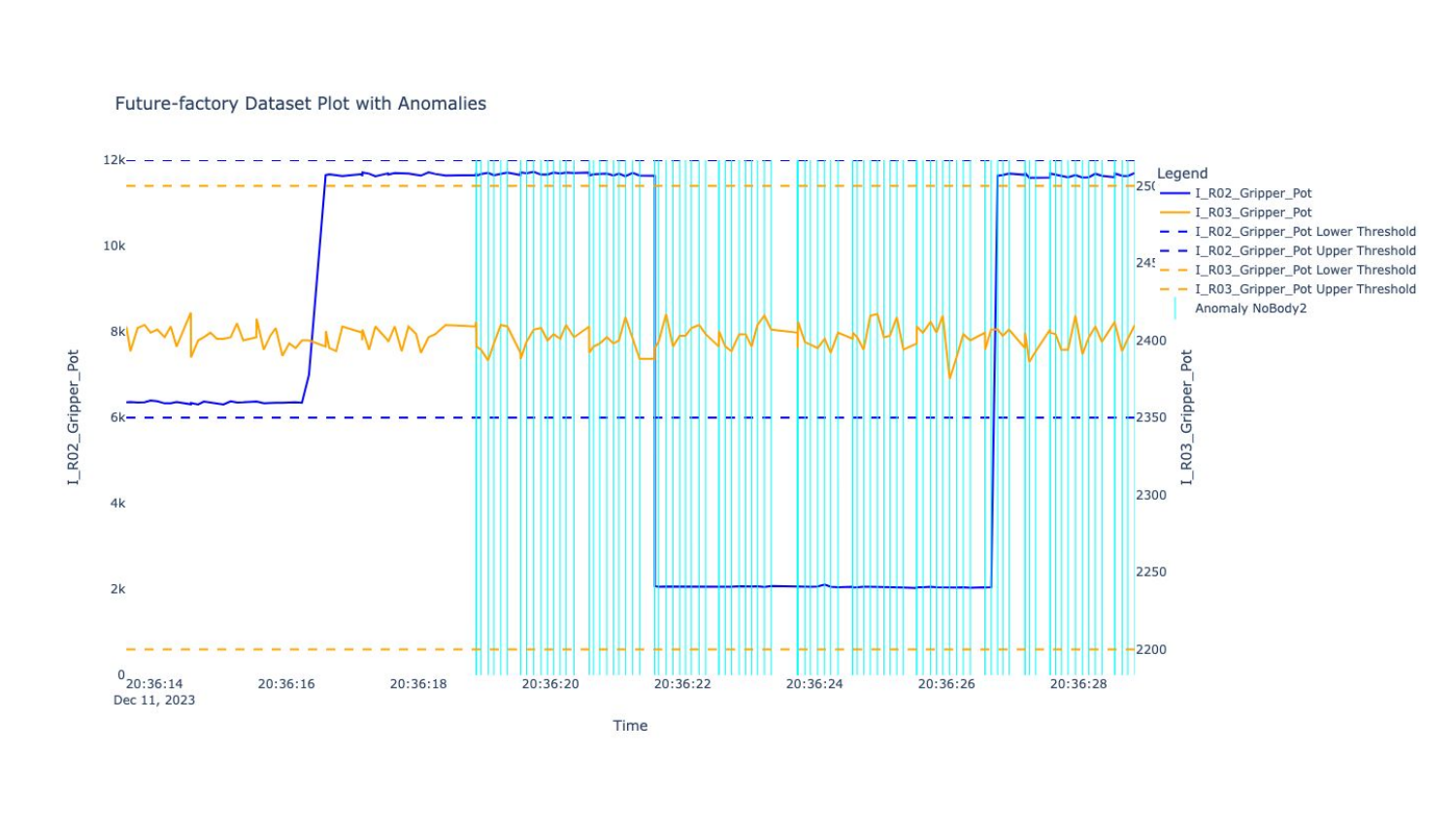}
    \caption{Time series plot for FF dataset with anomalies and thresholds for cycle states (cycle state 5,6,7) of a selected cell cycle: Anomaly Type- \textbf{NoBody2}. The dashed horizontal lines represent the upper and lower thresholds of the features. The vertical colored lines indicate the presence of an anomaly.}
    \label{fig:FF1}
  \end{subfigure}
  \hfill
  \begin{subfigure}[b]{0.45\textwidth}
    \includegraphics[width=\textwidth]{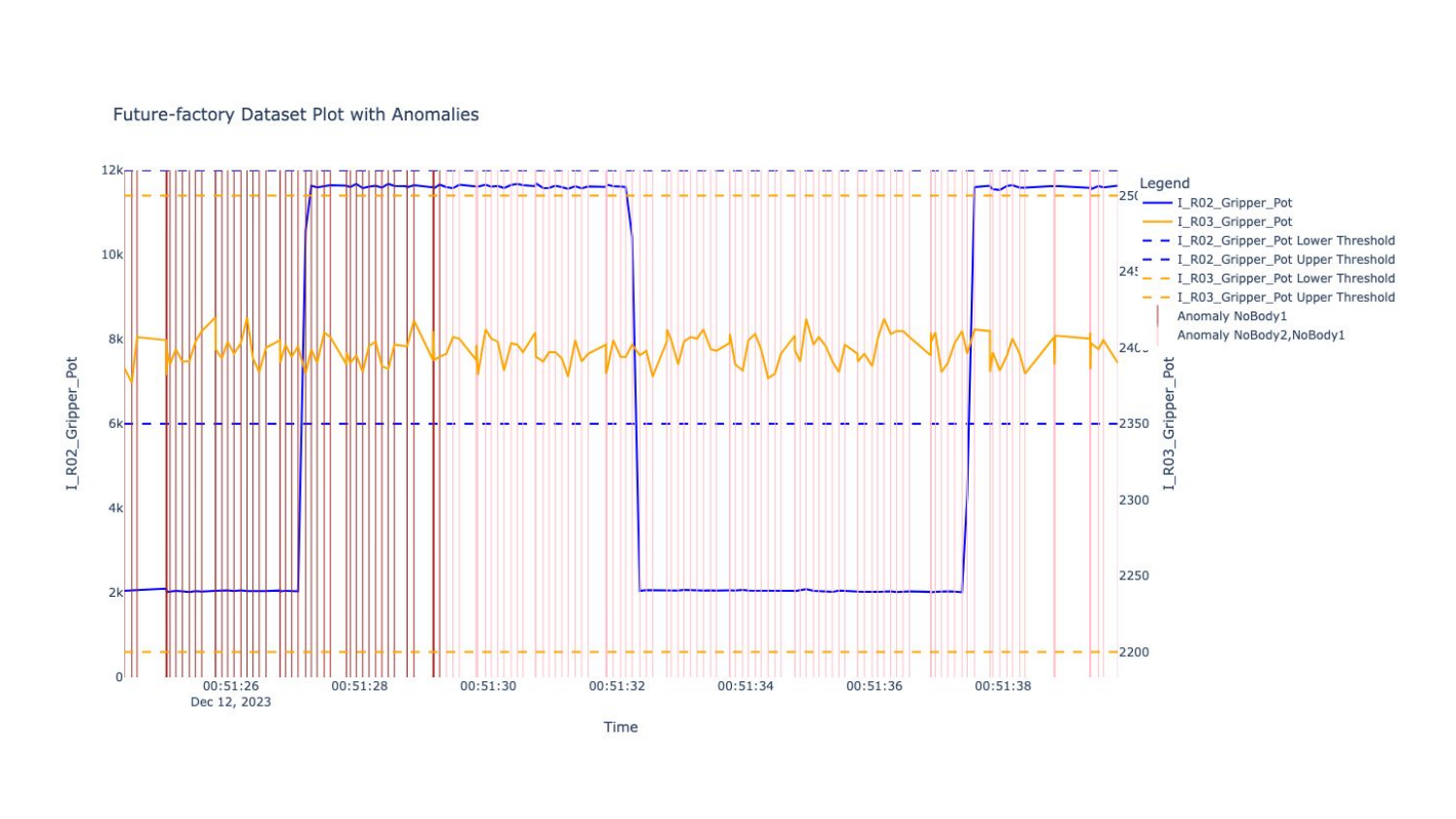}
    \caption{Time series plot for FF dataset with anomalies and thresholds for cycle states (cycle state 5,6,7) of a selected cell cycle: Anomaly Types- \textbf{NoBody1; NoBody2, NoBody1}. The dashed horizontal lines represent the upper and lower thresholds of the features. The vertical colored lines indicate the presence of an anomaly.}
    \label{fig:FF2}
  \end{subfigure}

  \caption{Selected EDA plots for Pulp and Paper Manufacturing Dataset and FF Dataset.}
  \label{fig:combined_eda}
\end{figure}

This section includes selected snapshots of the EDA conducted. Figure \ref{fig:x1} and figure \ref{fig:x2} illustrates EDA plots for Pulp and paper manufacturing dataset, whereas figure \ref{fig:FF5}, figure \ref{fig:FF6}, figure \ref{fig:FF1} and figure \ref{fig:FF2} are the plots of the FF dataset respectively.  

\clearpage

\subsection{Appendix C. Experimental Details on the Reference Time Series Foundational Models}
\label{sec:appendixA2}

\begin{enumerate}
    \item \textbf{FPT (Frozen Pre-trained Transformer)}
    \begin{itemize}
 \item FPT model is capable of performing both next-time series forecasting and anomaly detection tasks. 

\item Initially, we experiment with anomaly detection task with the MSL dataset. The preprocessed MSL dataset has three separate files; one containing the training features, one with the test features and one with the test labels. Anomaly detection is experimented on this preprocessed MSL dataset.
Then, we experiment with forecasting tasks with the ETTh1 dataset. 
 
 \item For the Future Factories dataset, we select the top 20 statistically significant features. 
 To fit the requirements of the FPT model for anomaly detection, we transformed the raw FF dataset into three separate files: one containing the training features, one with the test features and one with the test labels. The dataset is split into training, validation and testing sets with an 70/20/10 ratio. For anomaly detection, we convert multi-class labels to binary labels, as the FPT model only supports multi-featured binary-labeled data. For forecasting, we employ "LongTermForecasting" technique, which assess performance by averaging metric-results over prediction lengths of 96, 192, 336, and 720.  We kept the sampling frequency as 100 ms for the dataset.

 \item We use the FPT model for anomaly detection and time series forecasting on the pulp and paper dataset with the 59 features included (excluding categorical variables x28 and x61). To adapt the raw pulp and paper dataset for the FPT model for anomaly detection, we preprocessed it into three distinct files: one for training features, one for test features and one for test labels.
 The train, validation, and test split was also 70/20/10. We selected a two-minute sampling frequency for the dataset to match its original frequency of two minutes. For forecasting, the performance metrics are averaged across prediction lengths of 96, 192, 336, and 720.

  \item We selected a subset of the SMD dataset, consisting of 75,877 rows, for anomaly detection and time series forecasting using the FPT model. The dataset, with 37 features, was preprocessed into three separate files: one for training features, one for test features, and one for test labels, to suit the FPT model. The data was split into training, validation, and test sets with a ratio of 70/20/10, while maintaining the original frequency of the dataset for the experiments. In the forecasting task, we average the performance metric results over prediction lengths of 96, 192, 336, and 720.

\item Similarly, for the ETTh1 dataset, we use the FPT model for anomaly detection and time series forecasting on the SMD dataset with the 7 features included. To adapt the raw ETTh1 dataset for the FPT model for anomaly detection, we preprocessed it into three distinct files: one for training features, one for test features and one for test labels. The train, validation, and test split was also 70/20/10. We selected a one-hour sampling frequency for the dataset to match its original frequency of one-hour. For forecasting, the performance metrics are averaged across prediction lengths of 96, 192, 336, and 720.

\end{itemize}
    
    \item \textbf{TimeGPT}
\begin{itemize}
\item TimeGPT is capable of performing both anomaly detection (zero-shot) and next-time series forecasting (with fine-tuning on the training dataset). We aimed to evaluate the framework's capability on our datasets. 

\item For the Future Factory dataset, we chose a sampling frequency of one minute. Upon analyzing the data, we found that crucial information was retained at this frequency. This approach is primarily to optimize computational efficiency and prevent the TimeGPT API from crashing due to overly granular data. For time series forecasting, we split our data into an 80/20 train-test ratio. We select the features deemed important by domain experts [Features: I\_R02\_Gripper\_Pot, I\_R03\_Gripper\_Pot, I\_R03\_Gripper\_Load]. Our experiments show that the best results were achieved using these three features.  For anomaly detection, we convert multi-class labels to binary labels, as the TimeGPT model only supports multi-featured binary labeled data (all the anomalies are being labeled as one and non-anomalous states are labeled as zero, reference to the kind of anomaly is lost).

\item For the pulp dataset, we retained the existing time frequency of two minutes and included all 59 features for evaluation, excluding categorical variables x28 and x61. The data was split into an 80/20 train-test ratio for time series forecasting.

We then compare the results from these two experiments to analyze how varying the confidence interval affects anomaly detection.

\item
For the SMD, MSL, and ETTh1 datasets, we retain the original frequencies and utilize all the features from the datasets for both anomaly detection and prediction tasks.

\end{itemize}

\item \textbf{Time-MOE}
\begin{itemize}
\item 
Time-MOE is designed to perform next-step time series forecasting through fine-tuning on the training dataset. For the five datasets used in this study, we followed a consistent experimental procedure, demonstrating that the model exhibits strong generalization capabilities across different datasets.
\item 
We utilize the TimeMoE-50M model for time series forecasting. Initially, a tensor of random values (seqs) is generated to represent the input sequences, with a batch size of 2 and a context length of 12. The model is then loaded from the pre-trained Maple728/TimeMoE-50M checkpoint. The input sequences are normalized by subtracting the mean and dividing by the standard deviation along the last dimension. For forecasting, the model generates new tokens (predictions) based on the normalized input sequences. The predicted values are extracted and subsequently inverse normalized by multiplying by the original standard deviation and adding the original mean, resulting in the final forecasted values. This process is repeated for different prediction lengths (96, 192, 336, and 720), with the average performance metric calculated across these varying lengths.

\end{itemize}

\item \textbf{MOIRAI}
\begin{itemize}
    \item MOIRAI is primarily designed as a forecasting model and lacks built-in anomaly detection capabilities. Its main application is predicting future values, rather than identifying anomalies in real-time. Hence, it is utilized for the anomaly prediction task.
    \item In this study, we use its pretrained small model variant, fine-tuned for anomaly detection on various datasets. Predictions are generated with a context length of 200, batch size of 16, and 80\% confidence intervals for evaluation.
    \item For the Future Factories (FF) dataset, we resampled the data to a 1-second frequency to capture more granular variations. The anomaly label column was treated as the target variable, while the other features were used as covariates. The model was trained to forecast potential anomalies based on past data, leveraging its forecasting mechanism to identify potential future anomaly occurrences.
    \item In the Pulp dataset, the data was resampled to a 2-minute frequency to match the dataset's original frequency. Similar to the FF dataset, the anomaly label column was the target for forecasting, with the other 59 features acting as covariates. MOIRAI was then tasked with forecasting anomalies in the future rather than detecting them at the point of occurrence.
    \item For the SMD, MSL, and ETTh1 datasets, we used their respective granularities and feature sets without resampling.

    \item MOIRAI performs anomaly forecasting by training on the historical data of these datasets, but it does not detect anomalies as they occur in real-time, which differentiates it from models like TimeGPT or FPT.
\end{itemize}

\item \textbf{Chronos}
\begin{itemize}
\item Chronos is designed for next-time series prediction and currently lacks a mechanism for anomaly detection. Therefore, it is utilized for the anomaly prediction task. We use the smallest variant, Chronos-tiny (8M parameters), for our experiments, fine-tuned for forecasting on approximately 42 publicly available time-series datasets and their synthetic variants.
\item Predictions are generated with a context length of 512 tokens and a batch size of 64. The input sequences are tokenized by dividing by the mean of the time series entries in a particular context window and then quantizing those values into uniform bins before feeding them into the model. During prediction, the values are unscaled and dequantized. Model performance is assessed on a prediction length of 24, using the 0.5 quantile of the probabilistic-forecast obtained, with the rest of the dataset used for fine-tuning the model. While calculating the RMSE and MSE scores, we compared the normalized scaled values rather than the absolute ones, as they are sensitive to outliers (as we are using features of different scales).

\item For the Future Factories dataset, we resampled the data to a frequency of 1 minute. We selected the features deemed important by domain experts [Features: I\_R02\_Gripper\_Pot, I\_R03\_Gripper\_Pot, I\_R03\_Gripper\_Load]. Our experiments show that the best results were achieved using these three features.

\item For the Pulp dataset, we resampled the data to a frequency of 2 minutes, discarding the categorical variables x28 and x61, and including the remaining 59 features for evaluation.

\item For the SMD, MSL, and ETTh1 datasets, we retained the original frequencies and utilized all the features from the datasets for both anomaly detection and prediction tasks.

\item Chronos can be utilized for anomaly prediction in two ways. The first method, which we used here, involves using the forecasting pathway and then identifying anomalies by comparing the forecasted values with predefined safe thresholds. The second method involves extracting embeddings (representations) from the encoder layer and using these embeddings as input to a classifier to identify anomalies, provided we have labeled data. This is an approach we would like to explore further.

 \end{itemize}

\clearpage

\end{enumerate}

\subsection{Appendix D. Getting to Anomaly Prediction  from the Forecasted Values}
To obtain the results of anomaly prediction using time series forecasting of various features, it is essential to have predefined safe operating thresholds for each feature under specific operational conditions. These thresholds allow for the accurate classification of anomalies. For instance, consider three features: Feature 1, Feature 2, and Feature 3. By comparing the forecasted values of these features against their respective thresholds, we can classify each feature as either safe or unsafe. If the forecasted value of any feature falls into the unsafe zone, the presence of an outlier can be marked as true.

\subsection{Appendix E. Evaluation Results of the Analysis of Computational Cost of FMs with Statistical and Deep Learning Models}
\begin{table*}[!ht]
\scriptsize
\centering
\label{tab:results_transposed}
\begin{tabular}{c|c|c|c|c} 
\hline
Dataset & \#Features & TSFM           & \begin{tabular}[c]{@{}c@{}}Zero-Shot Anomaly \\Detection Time (min)\end{tabular} & \begin{tabular}[c]{@{}c@{}}Zero-Shot Forecasting Time \\{[}Anomaly Prediction] (min)\end{tabular}  \\ 
\hline
FF      & 2          & TimeGPT        & 6.1                                                                              & 8.2                                                                                                \\
        & 20         & FPT     & 30                                                                               & 43                                                                                                 \\
        & 20         & Time-MOE(base) & -                                                                                & 2                                                                                                  \\
        & 20         & MOIRAI(base)   & -                                                                                & 3                                                                                                  \\
        & 20         & Chronos(tiny)  & -                                                                                & 1                                                                                                  \\
        & 20         & Autoencoder    & 0.84                                                                             & 0.91                                                                                               \\
        & 20         & Weighted XGB   & 0.6                                                                              & 0.61                                                                                               \\ 
\hline
Pulp    & 59         & TimeGPT        & 10.6                                                                             & 14.5                                                                                               \\
        & 59         & FPT     & 37                                                                               & 48                                                                                                 \\
        & 20         & Time-MOE(base) & -                                                                                & 3.6                                                                                                \\
        & 20         & MOIRAI(base)   & -                                                                                & 3.8                                                                                                \\
        & 20         & Chronos(tiny)  & -                                                                                & 1.1                                                                                                \\
        & 20         & Autoencoder    & 0.85                                                                             & 0.87                                                                                               \\
        & 20         & Weighted XGB   & 0.57                                                                             & 0.6                                                                                                \\ 
\hline
SMD     & 37         & TimeGPT        & 15.1                                                                             & 21.1                                                                                               \\
        & 37         & FPT     & 34                                                                               & 49                                                                                                 \\
        & 37         & Time-MOE(base) & -                                                                                & 2.1                                                                                                \\
        & 37         & MOIRAI(base)   & -                                                                                & 4.9                                                                                                \\
        & 37         & Chronos(tiny)  & -                                                                                & 1                                                                                                  \\
        & 37         & Autoencoder    & 0.81                                                                             & 0.95                                                                                               \\
        & 37         & Weighted XGB   & 0.63                                                                             & 0.69                                                                                               \\ 
\hline
MSL     & 55         & TimeGPT        & 8.9                                                                              & 8.2                                                                                                \\
        & 55         & FPT     & 30                                                                               & 43                                                                                                 \\
        & 55         & Time-MOE(base) & -                                                                                & 8.1                                                                                                \\
        & 55         & MOIRAI(base)   & -                                                                                & 3                                                                                                  \\
        & 55         & Chronos(tiny)  & -                                                                                & 1                                                                                                  \\
        & 55         & Autoencoder    & 0.8                                                                              & 0.88                                                                                               \\
        & 55         & Weighted XGB   & 0.7                                                                              & 0.72                                                                                               \\ 
\hline
ETTh1   & 7          & TimeGPT        & 12.3                                                                             & 24.7                                                                                               \\
        & 7          & FPT     & 20.6                                                                             & 31.8                                                                                               \\
        & 7          & Time-MOE(base) & -                                                                                & 43                                                                                                 \\
        & 7          & MOIRAI(base)   & -                                                                                & 2                                                                                                  \\
        & 7          & Chronos(tiny)  & -                                                                                & 3                                                                                                  \\
        & 7          & Autoencoder    & 2                                                                                & 1                                                                                                  \\
        & 7          & Weighted XGB   & 0.54                                                                             & 0.6                                                                                                \\
\hline
\end{tabular}
\caption{Inference Time taken for Zero-Shot Anomaly Detection and Prediction (min)}
\end{table*}
\clearpage


\end{document}